\documentclass{article}
\usepackage{spconf,amsmath,graphicx}
\usepackage{mathtools}
\usepackage{units}

\title{A Novel Focal Tversky loss function with Improved Attention U-Net for lesion segmentation}
\name{Nabila Abraham, Naimul Mefraz Khan
\thanks{Code available at https://github.com/nabsabraham/focal-tversky-unet}}
\address{Ryerson University \\ 
Department of Electrical and Computer Engineering \\
350 Victoria Street, Toronto, ON}

\begin{document}
%
\maketitle
\begin{abstract}
We propose a generalized focal loss function based on the Tversky index to address the issue of data imbalance in medical image segmentation. Compared to the commonly used Dice loss, our loss function achieves a better trade off between precision and recall when training on small structures such as lesions. To evaluate our loss function, we improve the attention U-Net model by incorporating an image pyramid to preserve contextual features. We experiment on the BUS 2017 dataset and ISIC 2018 dataset where lesions occupy 4.84\% and 21.4\% of the images area and improve segmentation accuracy when compared to the standard U-Net by 25.7\% and 3.6\%, respectively.  
\end{abstract}


\section{Introduction}
\label{sec:intro}

A common task in medical image analysis is the ability to detect and segment pathological regions that typically occupy a very small fraction of the full image. Such imbalance in the data can lead to instability in established generative and discriminative frameworks \cite{sudre}. In recent literature, convolutional neural networks (CNNs) have been successfully applied to automatically segment 2D and 3D biological data \cite{sudre}. Most of the current deep learning methods derive from a fully convolutional network architecture (FCN), where the fully connected layers are replaced by convolutional layers \cite{long}. The popular U-Net is an FCN variant which has become the defacto standard for image segmentation due to its multi-scale skip connections and learnable up-convolution layers \cite{ronne}. 

A dominant research area in image segmentation is to develop strategies to deal with class imbalance. The focal loss function proposed in \cite{focalloss} reshapes the cross-entropy loss function with a modulating exponent to down-weight errors assigned to well-classified examples. The focal loss prevents the vast number of easy negative examples from dominating the gradient to alleviate class-imbalance. In practice however, it faces difficulty balancing precision and recall due to small regions-of-interest (ROI) found in medical images. Research efforts to address small ROI segmentation propose more discriminative models such as attention gated networks \cite{oktay}, \cite{lin}. CNNs with attention gates (AGs) focus on the target region, with respect to the classification goal, and can be trained end-to-end. At test time, these gates generate soft region proposals to highlight salient ROI features and suppress feature activations by irrelevant regions. 

To address the issues of data imbalance and training performance, we combine attention gated U-Net with a novel variant of the focal loss function, better suited for small lesion segmentation. Our major contributions include (1) a novel focal Tversky loss function for highly imbalanced data and small ROI segmentation, where we modulate the Tversky index \cite{tversky} to improve precision and recall balance, and (2) a deeply supervised attention U-Net \cite{oktay}, improved with a multi-scaled input image pyramid for better intermediate feature representations. Experiments were performed on the Breast Ultrasound Lesions 2017 dataset B (BUS) \cite{yap} and the ISIC 2018 skin lesion dataset \cite{isic1}, \cite{isic2}, both datasets suffering from class imbalance and large intra class variation in lesion sizes. On average, the lesions occupy 4.84\% $\pm$ 5.43\% and 21.4\% $\pm$ 20.3\% in ISIC 2018 dataset and BUS dataset B, respectively. When compared to the baseline U-Net, our methods improves Dice scores by 25.7\% and 3.6\% for BUS dataset B and ISIC 2018, respectively.

\section{Methodology}

\subsection{Focal Tversky Loss}

\begin{figure}[h]
	\centering
	\includegraphics[width=6cm]{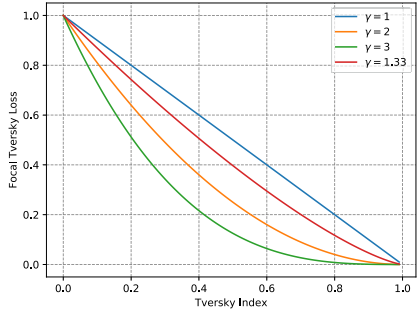}	
	\caption{The focal Tversky loss non-linearly focuses training on hard examples (where Tversky Index $<0.5$) and suppresses easy examples from contributing to the loss function. }
	\label{fig:focal}
\end{figure}    

In the medical community, the Dice score coefficient (DSC) is an overlap index that is widely used to asses segmentation maps. The 2-class DSC variant for class $c$ is expressed in Equation \ref{eq:dsc}, where $g_{ic} \in$ $\{0,1\}$  and $p_{ic} \in [0,1]$ represent the ground truth label and the predicted label, respectively. The total number of pixels in an image is denoted by $N$. The $\epsilon$ provides numerical stability to prevent division by zero. 

\begin{equation}
  DSC_c = \frac{\sum_{i=1} ^N p_{ic} g_{ic} + \epsilon}{\sum_{i=1} ^N p_{ic} + g_{ic} + \epsilon} 
\label{eq:dsc} 
\end{equation}

A common method to reduce the effects of class imbalance is to introduce a weight $w_c$ for each class $c$, which is inversely proportional to the label frequency \cite{wong}.  The linear Dice loss (DL) is therefore defined as a minimization of the overlap between the prediction and ground truth \cite{milletari}: 
\begin{equation}
  DL_c = \sum_{c} 1 - DSC_c
\label{eq:DL} 
\end{equation}

One of the limitations of the Dice loss function is that it equally weighs false positive (FP) and false negative (FN) detections. In practice, this results in segmentation maps with high precision but low recall. With highly imbalanced data and small ROIs such as skin lesions, FN detections need to be weighted higher than FPs to improve recall rate. The Tversky similarity index is a generalization of the Dice score which allows for flexibility in balancing FP and FNs:

\begin{equation}
  TI_c = \frac{\sum_{i=1} ^N p_{ic} g_{ic} + \epsilon}{\sum_{i=1} ^N p_{ic}g_{ic} + \alpha\sum_{i=1} ^N p_{i\bar{c}}g_{ic} + \beta\sum_{i=1} ^N p_{ic}g_{i\bar{c}} + \epsilon} 
\label{eq:tversky} 
\end{equation}

where, $p_{ic}$ is the probability that pixel $i$ is of the lesion class $c$ and $p_{i\bar{c}}$ is the probability pixel $i$ is of the non-lesion class, $\bar{c}$. The same is true for $g_{ic}$ and $g_{i\bar{c}}$, respectively. Hyperparameters $\alpha$ and $\beta$ can be tuned to shift the emphasis to improve recall in the case of large class imbalance. The Tversky index is adapted to a loss function (TL) in \cite{tversky} by minimizing  $\sum_{c} 1 - TI_c$. 

Another issue with the DL is that it struggles to segment small ROIs as they do not contribute to the loss significantly. To address this, we propose the focal Tversky loss function (FTL), parametrized by $\gamma$, for control between easy background and hard ROI training examples. In \cite{focalloss}, the focal parameter exponentiates the cross-entropy loss to focus on hard classes detected with lower probability. This idea has been extended in recent works where an exponent is applied to the Dice score \cite{wang} or a combination of Dice and cross-entropy \cite{wong}, \cite{zhu}. Similarly, we define our Focal Tversky Loss (FTL) function as:
 
\begin{equation}
  FTL_c =  \sum_{c} (1 - TI_c)^{\nicefrac{1}{\gamma}}
\label{eq:FTL} 
\end{equation}

where $\gamma$ varies in the range $[1,3]$. In practice, if a pixel is misclassified with a high Tversky index, the FTL is unaffected. However, if the Tversky index is small and the pixel is misclassified, the FTL will decrease significantly. 

When $\gamma>1$, the loss function focuses more on less accurate predictions that have been misclassified. However, we observe over-suppression of the FTL when the class accuracy is high, usually as the model is close to convergence. This trend is visualized in Figure \ref{fig:focal} as increasing values of Tversky index are mapped to flatter regions of the FTL curve with increasing values of $\gamma$. We experiment with high values of $\gamma$ and observe the best performance with $\gamma = \frac{4}{3}$ and therefore train all experiments with it. To combat the over-suppression of the loss function, we train intermediate layers with the FTL but supervise the last layer with the Tversky loss to provide a strong error signal and mitigate sub-optimal convergence. 

We hypothesize using a higher $\alpha$ in our generalized loss function will improve model convergence by shifting the focus to minimize FN predictions. Therefore, we train all models with $\alpha=0.7$ and $\beta=0.3$. It is important to note that in the case of $\alpha$ = $\beta$ = 0.5, the Tversky index simplifies to the DSC. Moreover, when $\gamma = 1$, the FTL simplifies to the TL.

\subsection{Network Architecture}

\begin{figure*}[h]
	\centering
	\includegraphics[width=15cm]{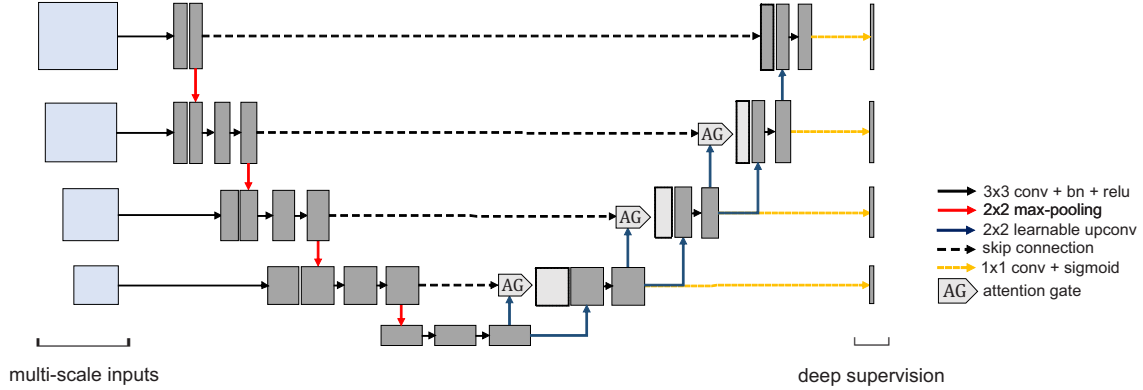}
	\caption{Proposed Attention U-Net architecture with input image pyramid and deep supervised output layers. }
	\label{fig:net}
\end{figure*}

To achieve further balance between precision and recall, we propose an improved attention U-Net \cite{oktay} that incorporates the proposed FTL. This architecture is based on the popular U-Net which has been designed to work well with very small number of training examples (Figure \ref{fig:net}). The network is composed of a contracting path to extract locality features and an expansive path, to resample the image maps with contextual information. Skip connections are used to combine high-resolution local features with low-resolution global features and encourage more semantically meaningful outputs.

At the deepest stage of encoding, the network has the richest possible feature representation. However, with cascaded convolutions and non-linearities, spatial details tend to get lost in the high-level output maps. This makes it difficult to reduce false detections for small objects that show large shape variability \cite{oktay}. To address this issue, we use soft attention gates (AGs) to identify relevant spatial information from low-level feature maps and propagate it to the decoding stage.

\begin{figure}
	\centering
	\centerline{\includegraphics[width=6.5cm]{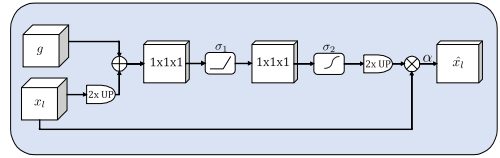}}
	\caption{ Schematic of additive attention gate (AG) adapted from 			\cite{oktay}. Input features $x_l$ are scaled with attention coefficients $\alpha_i$ to propagate relevant features to the decoding layer output $\hat{x_l}$. The coarser gating signal $g$ provides contextual information while spatial regions from the input $x_l$ provide locality information. Feature map resampling is computed by bilinear interpolation. }
\label{fig:ag}	
\end{figure}

AGs produce attention coefficients $\alpha_i \in [0,1]$ at each pixel $i$, that scale input feature maps $x^l_i$, at layer $l$, to output semantically relevant features, $\hat{x_i}^l$ , as depicted in Figure \ref{fig:ag}. A gating signal, $g$, is used for each pixel $i$ to determine focus regions. It is collected from a coarser scale than the input query signal, $x^l_i$ to compute intermediate activation maps: 
 
\begin{equation}
q^l_{attn} = \psi^T ( \sigma_1 ( W_x^T x_i^l + W_g^T g_i + b_g)) + b_{\psi}
\label{eq:qattn}
\end{equation}

where the linear attention coefficients, $q^l_{attn}$, are computed by the element-wise sum and 1x1 linear transformations, parameterized by $W_x$, $b_x$, $W_g$ and $b_g$.  The intermediate maps are transformed by ReLU and sigmoid non-linearities applied as $\sigma_1$ and $\sigma_2$, respectively:  

\begin{equation}
\alpha_i^l = \sigma_2 ( q^l_{attn}(x_i^l, g_i) 
\label{eq:alpha}
\end{equation} 

The attention coefficients $\alpha_i$ scale the low level query signal $x^l_i$ by an element-wise product and retain only relevant activations. These pruned features are then concatenated with upsampled output maps at each scale in the expansive stage. The lowest-level feature maps, i.e. the first skip connections, are not used in the gating function as they do not represent input data in a high dimensional space \cite{oktay}. A 1x1x1 convolution and sigmoid activation is applied on each output map in the expansive stage. Every high dimensional feature representation is supervised with our FTL, with the exception of the last layer, to avoid loss over-suppression. This tactic of deep supervision, introduced in \cite{lee}, forces intermediate layers to be semantically discriminative at every scale. Moreover, it helps to ensure that attention unit has the ability to influence the responses to a large range of image foreground content. 

Moreover, since different kinds of class details are more
easily accessible at different scales, we inject the encoder layers with an input image pyramid before each of the max-pooling layers. Combined with deep supervision, this method improves segmentation accuracy for datasets where small ROI features can get lost in cascading convolutions and facilitates the network learning more locality aware features with respect to the classification goal.

\section{Experiments}
\label{sec:results}

\begin{table*}[h!]
\caption{Performance on BUS 2017 Dataset B with 40 test images\strut} 

\centering 
\begin{tabular}{c c c c c} 

\hline \hline 
Model & Parameters & DSC  & Precision & Recall \\[0.5ex] 
\hline 

U-Net + DL & $\alpha=0.5$, $\beta=0.5$ & 0.547 $\pm$ 0.04 & 0.653 $\pm$ 0.171 & 0.658 $\pm$ 0.146\\ 
U-Net + TL & $\alpha=0.7$, $\beta=0.3$ & 0.657 $\pm$ 0.02 & 0.732 $\pm$ 0.072 & 0.723 $\pm$ 0.074 \\
U-Net + FTL & $\alpha=0.7$, $\beta=0.3$, $\gamma=\nicefrac{4}{3}$ & 0.669 $\pm$ 0.033 & 0.775 $\pm$ 0.047 & 0.715 $\pm$ 0.057 \\
Attn U-Net + DL & $\alpha=0.5$, $\beta=0.5$ & 0.615 $\pm$ 0.020 & 0.675 $\pm$ 0.042 & 0.658 $\pm$ 0.049 \\
Attn U-Net + Multi-Input + DL & $\alpha=0.5$, $\beta=0.5$ & 0.716 $\pm$ 0.041 & 0.759 $\pm$ 0.092 & 0.751 $\pm$ 0.046 \\
Attn U-Net + Multi-Input + TL & $\alpha=0.7$, $\beta=0.3$ & 0.751 $\pm$ 0.042 & 0.802 $\pm$ 0.073 & 0.768 $\pm$ 0.056 \\
Attn U-Net + Multi-Input + FTL & $\alpha=0.7$, $\beta=0.3$, $\gamma=\nicefrac{4}{3}$ & \textbf{0.804} $\pm$ \textbf{0.024} & \textbf{0.829} $\pm$ \textbf{0.027} & \textbf{0.817} $\pm$ \textbf{0.022} \\

\end{tabular}
\label{table:bus} 
\end{table*}

\begin{table*}[h!]
\caption{Performance on ISIC 2018 with 649 test images \strut} 

\centering 
\begin{tabular}{c c c c c} 

\hline \hline 
Model & Parameters & DSC  & Precision & Recall \\[0.5ex] 
\hline 

U-Net + DL & $\alpha=0.5$, $\beta=0.5$ & 0.820 $\pm$ 0.013 & 0.849	 $\pm$ 0.038 & 0.867 $\pm$ 0.048  \\ 
U-Net + TL & $\alpha=0.7$, $\beta=0.3$ & 0.838 $\pm$ 0.026 & 0.822 $\pm$ 0.051 & 0.917 $\pm$ 0.033\\
U-Net + FTL & $\alpha=0.7$, $\beta=0.3$, $\gamma=\nicefrac{4}{3}$ & 0.829	 $\pm$ 0.027 & 0.797 $\pm$ 0.040 & \textbf{0.926 $\pm$ 0.012}\\
Attn U-Net + DL & $\alpha=0.5$, $\beta=0.5$ & 0.806 $\pm$ 0.033 & 0.874 $\pm$ 0.080 & 0.827 $\pm$ 0.055 \\
Attn U-Net + Multi-Input + DL & $\alpha=0.5$, $\beta=0.5$ & 0.827 $\pm$ 0.055 & \textbf{0.896 $\pm$ 0.019} & 0.829 $\pm$ 0.076  \\
Attn U-Net + Multi-Input + TL & $\alpha=0.7$, $\beta=0.3$ & 0.841 $\pm$ 0.012 & 0.823 $\pm$ 0.038 & 0.912 $\pm$ 0.026 \\
Attn U-Net + Multi-Input + FTL & $\alpha=0.7$, $\beta=0.3$, $\gamma=\nicefrac{4}{3}$ & \textbf{0.856 $\pm$ 0.007} & 0.858	 $\pm$ 0.020 & 0.897 $\pm$ 0.014 \\

\end{tabular}
\label{table:isic} 
\end{table*}

We validate the FTL on two datasets where the ROI class is significantly smaller than the background class and observe large performance gains. We experiment with the Breast Ultrasound Dataset B (BUS) open-sourced in \cite{yap}. This dataset consists of 163 ultrasound images of breast lesions from different women. The average image size is 760 x 570 pixels where each of the images presented one or more lesions. For our experiments, the data is resampled to 128 x 128 pixels with a 75-25 train-test split. To extend our proposed method to larger datasets, we extract training data from the ISIC 2018: Skin Lesion Analysis Towards Melanoma Detection” grand challenge dataset \cite{isic1}, \cite{isic2}. This dataset consists of 2,594 RGB images of skin lesions with an average image size of 2166 x 3188 pixels. For our experiments, the dataset is resampled to 192 x 256 pixels with 75-25 train-test split.

To present a fair evaluation of our multi-scaled attention U-Net and the focal Tversky loss, we do not augment our datasets or incorporate any transfer learning. We study 7 cases of variations within U-Net and the Tversky loss function while comparing to the baseline U-Net trained with Dice loss. Ablation test results are recorded in Section \ref{sec:results} with 5-fold cross validation for Dice scores, precision and recall. 

The ISIC 2018 experiment was trained for 50 epochs with a batch size of 8. The BUS 2017 dataset was trained for 100 epochs with a batch size of 16. Both models were optimized using stochastic gradient descent with momentum, using an initial learning rate at 0.01 which decays by $10^{-6}$ on every epoch. These parameters were optimized through a grid search method \cite{gridsearch}. All experiments are programmed using the Keras framework with the Tensorflow backend and trained using an NVIDIA GTX 1070 GPU. Open-source implementation with reproducible results for this paper can be obtained from https://github.com/nabsabraham/focal-tversky-unet.

\section{Results}
Table \ref{table:bus} shows that the baseline U-Net trained with the Dice loss function has the worst performance. The large standard deviation in the precision and recall scores suggest the learning is not stable. In contrast, U-Net models trained with TL and FTL show increased DSC and more balanced precision-recall scores which occurs due to weighting $\alpha$ higher in the loss function than $\beta$. We observe incorporating attention in U-Net trained with DL depicts lower Dice scores than the baseline, probably due to the intra-lesion variation. Injecting an input pyramid into the model improves the DSC significantly suggesting features of small lesions are easily lost when class imbalance is high. Training the attention model with FTL combines the benefits of improved feature selection with focused training to outperform all other methods. The proposed architecture (last row) is able to segment lesions with a Dice score of $0.804$ on training with a small subset of 100 images. 

Contrary to the BUS scores, ISIC results in Table \ref{table:isic} show the baseline U-Net trained with DL performs well due to the large training sample size, variation in lesion structures and distinct features present in the RGB images. Training U-Net with TL and FTL, we observe an improved DSC score. However, when the Tversky index is high for misclassified examples, the focal exponent $\gamma$ suppresses the contribution to the error signal and since $\alpha$ is weighted higher than $\beta$, the model converges to the highest reported recall at $0.926$, but lowest precision. To address this issue, when training the proposed attention model, we supervise the last layer with TL so that a true error signal will still propogate back when the model is close to convergence. As a result, our improved attention U-Net model with FTL (last row) obtains slightly lower but overall better balanced recall and precision, and, consequently, the best DSC score. We outperform the baseline by $3.6\%$ with a low spread of $0.7\%$. 

\section{Conclusion}

In this work, we propose a novel focal Tversky loss function to improve the precision and recall balance in semantic segmentation. Our experiments demonstrate the importance of the choice of loss function when dealing with highly imbalanced problems and with varying dataset sizes. Moreover, we improve the attention U-Net proposed in \cite{oktay} by incorporating a an input image pyramid into each scale in the model architecture. The redundancy in the features helps recover any lost contextual information which is crucial when segmenting small ROIs such as lesions. Our proposed method outperforms the baseline U-Net in Dice scores and presents balanced precision-recall scores with low standard deviations.
\vfill
\pagebreak

\label{sec:ref}
\newpage
\bibliographystyle{IEEEbib}
\bibliography{strings,refs}

\end{document}